\documentclass[acmtog]{acmart}
\acmSubmissionID{1208}

\usepackage{multirow}
\usepackage{diagbox}

\usepackage{tikz}
\newcommand{\NAcell}{%
  \begin{tikzpicture}[baseline=(current bounding box.center)]
    \draw (0,0) -- (0.4,0.4);
  \end{tikzpicture}%
}

\author{Sylvia Yuan}
\affiliation{
  \institution{University of California San Diego}
  \country{USA}
}
\email{l1yuan@ucsd.edu}

\author{Ruoxi Shi}
\affiliation{
  \institution{University of California San Diego}
  \country{USA}
}
\affiliation{
  \institution{Hillbot Inc.}
  \country{USA}
}
\email{r8shi@ucsd.edu}

\author{Xinyue Wei}
\affiliation{
  \institution{University of California San Diego}
  \country{USA}
}
\affiliation{
  \institution{Hillbot Inc.}
  \country{USA}
}
\email{xiwei@ucsd.edu}

\author{Xiaoshuai Zhang}
\affiliation{
  \institution{Hillbot Inc.}
  \country{USA}
}
\email{x@hillbot.ai} 

\author{Hao Su}
\affiliation{
  \institution{University of California San Diego}
  \country{USA}
}
\affiliation{
  \institution{Hillbot Inc.}
  \country{USA}
}
\email{haosu@ucsd.edu}

\author{Minghua Liu}
\affiliation{
  \institution{Hillbot Inc.}
  \country{USA}
}
\email{m@hillbot.ai}

\copyrightyear{2025}
\acmYear{2025}
\setcopyright{acmlicensed}\acmConference[SA Conference Papers '25]{SIGGRAPH Asia 2025 Conference Papers}{December 15--18, 2025}{Hong Kong, Hong Kong}
\acmBooktitle{SIGGRAPH Asia 2025 Conference Papers (SA Conference Papers '25), December 15--18, 2025, Hong Kong, Hong Kong}
\acmDOI{10.1145/3757377.3763844}
\acmISBN{979-8-4007-2137-3/2025/12}


\begin{document}

\title{LARM: A Large Articulated‑Object Reconstruction Model }

\begin{teaserfigure}
  \includegraphics[width=\textwidth]{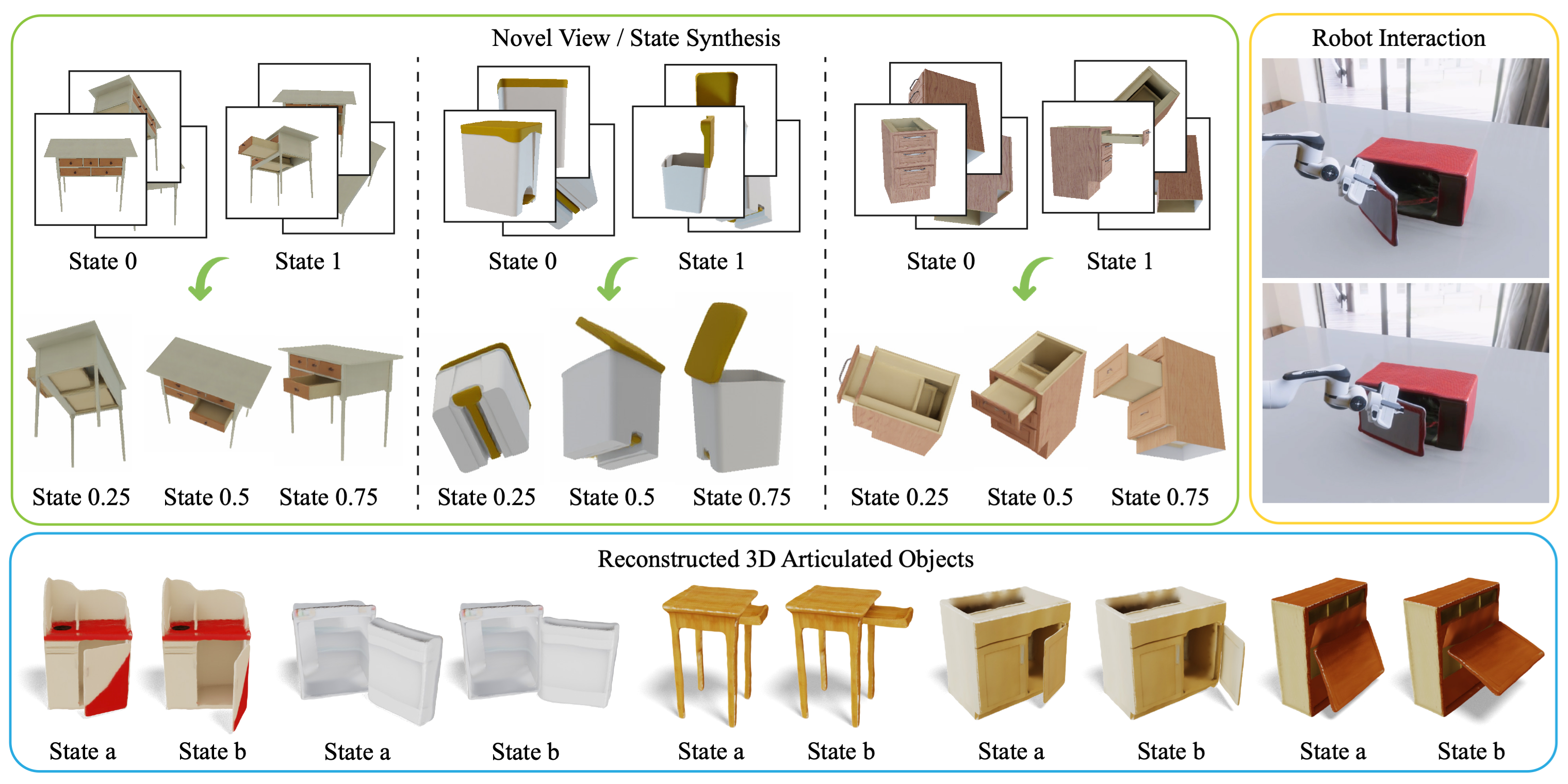}
  \vspace{-2.5em}
  \caption{We introduce LARM, a feedforward model for 3D articulated object reconstruction. Given sparse-view images of an articulated object in two different joint states (e.g., drawer open and closed), LARM can synthesize views from arbitrary camera poses and novel joint configurations (e.g., the drawer half-open), enabling efficient generation of continuous articulation and viewpoint variations.
Beyond novel view and state synthesis, LARM also supports explicit 3D mesh reconstruction. Unlike many recent methods that rely on part retrieval or template meshes—approaches that often fail to capture fine-grained geometry and lack support for realistic textures—LARM reconstructs high-quality, textured, articulated 3D objects that closely match the input images in terms of geometric detail, visual fidelity, and articulation accuracy. LARM enables a wide range of downstream applications, including high-fidelity digital twin reconstruction from casual real-world captures, which is crucial for large-scale robotics learning in simulation.}
  \label{fig:teaser}
\end{teaserfigure}

\begin{abstract}
Modeling 3D articulated objects with realistic geometry, textures, and kinematics is essential for a wide range of applications. However, existing optimization-based reconstruction methods often require dense multi-view inputs and expensive per-instance optimization, limiting their scalability. Recent feedforward approaches offer faster alternatives but frequently produce coarse geometry, lack texture reconstruction, and rely on brittle, complex multi-stage pipelines. We introduce LARM, a unified feedforward framework that reconstructs 3D articulated objects from sparse-view images by jointly recovering detailed geometry, realistic textures, and accurate joint structures. LARM extends LVSM—a recent novel view synthesis (NVS) approach for static 3D objects—into the articulated setting by jointly reasoning over camera pose and articulation variation using a transformer-based architecture, enabling scalable and accurate novel view synthesis. In addition, LARM generates auxiliary outputs such as depth maps and part masks to facilitate explicit 3D mesh extraction and joint estimation. Our pipeline eliminates the need for dense supervision and supports high-fidelity reconstruction across diverse object categories. Extensive experiments demonstrate that LARM outperforms state-of-the-art methods in both novel view and state synthesis as well as 3D articulated object reconstruction, generating high-quality meshes that closely adhere to the input images. Our project page is at \url{https://sylviayuan-sy.github.io/larm-site/}.
\vspace{-0.5em}
\end{abstract}

\maketitle

\vspace{-1.5em}
\section{Introduction}
\vspace{-0.3em}

3D articulated objects—from household appliances and furniture to complex mechanical assemblies—pervade everyday environments, yet faithfully modeling their geometry and kinematic structures remains a labor‑intensive endeavor. Traditional pipelines rely on skilled artists who manually build and rig each asset, incurring substantial time and monetary costs, and large‑scale datasets with detailed articulation labels are correspondingly scarce. Accurate reconstruction of such objects is, however, a prerequisite for high‑fidelity robotic simulation and digital‑twin systems, AR/VR and animation, where realistic geometry, appearance, and kinematics enable robust robotics training and low sim-to-real gap deployment. 

Prior work has made significant strides in analyzing articulated objects~\cite{liu2024survey}, particularly through tasks like joint axis prediction and joint state estimation~\cite{fu2024capt,liu2023category}. Beyond analysis, several optimization-based reconstruction methods—often built on NeRFs or 3D Gaussian Splatting—have been developed to recover articulated geometry~\cite{tseng2022cla, wei2022self, liu2023paris}. While effective, these approaches are computationally expensive, require dense multi-view inputs, and rely on per-instance optimization with limited shape-level priors, hindering scalability and generalization. More recently, feedforward models have emerged for articulated object reconstruction or generation from various inputs~\cite{le2024articulate,chen2024urdformer,liu2024singapo}. These input conditions include object categories, articulation graphs, single images, and even text. However, many of these models~\cite{le2024articulate,chen2024urdformer,liu2024singapo} oversimplify geometry with bounding boxes, template meshes, or retrieved parts from a small database, leading to coarse and unrealistic results. Many also lack texture reconstruction, further reducing visual fidelity. Consequently, they struggle to produce meshes that accurately reflect user inputs, limiting their applicability to use cases like digital twins and high-fidelity simulation. A few recent feedforward methods~\cite{zhang2021strobenet,kawana2023detection} model geometry with implicit fields (e.g., SDFs), but most of these methods involve complex, multi-stage pipelines, and still lack high-quality texture modeling.

In contrast, we propose LARM, a feedforward framework for reconstructing 3D articulated objects. Unlike prior methods, LARM jointly recovers detailed geometry, realistic textures, and accurate articulated structure from sparse-view inputs, eliminating the need for dense-view observations. Built on a simple and unified architecture, it is fast, scalable, and capable of learning rich shape priors across a wide variety of articulated objects. 

Our approach draws inspiration from LVSM~\cite{jin2024lvsm}, a recent method that demonstrates a transformer model with minimal 3D inductive bias can synthesize novel views of static objects. We extend this idea to articulated settings by enabling the model to reason over both camera pose and articulation variation. As shown in Figure~\ref{fig:teaser}, LARM takes as input sparse-view posed images taken from two articulation states (e.g., a rest state and a maximum state), each concatenated with its corresponding joint state. These inputs are processed by a transformer alongside target-view Plücker rays and a target joint state to predict the image tokens of the desired view. In doing so, LARM is trained to synthesize novel views conditioned on arbitrary camera poses and joint configurations, effectively capturing both viewpoint and articulation changes. While novel-view synthesis alone already supports a range of downstream applications, we further extend LARM to output auxiliary signals—including depth maps, foreground masks, and part segmentation masks—to facilitate explicit 3D reconstruction. Finally, we introduce post-processing modules that leverage these outputs for articulation joint estimation and explicit mesh extraction, enabling the full reconstruction of textured, articulated 3D assets.

We compare LARM with recent state-of-the-art methods for 3D articulated object reconstruction and generation, including both optimization-based and feedforward approaches. We evaluate performance on both novel view and state synthesis as well as 3D articulated object reconstruction tasks, where LARM significantly outperforms baseline methods across all tasks and metrics. Compared to baselines—especially retrieval-based ones—LARM demonstrates much better adherence to the input images in terms of both geometric and texture details. We also test our pipeline using casually captured images from handheld devices, showcasing the model’s potential for real-world applications such as AR/VR and large-scale digital twin generation for robot learning.
\vspace{-0.7em}
\section{Related Work}
\vspace{-0.5em}

\subsection{Understanding of 3D Articulated Objects}
\vspace{-0.5em}

Understanding the structure and motion of 3D articulated objects has been a central topic in both computer vision and robotics. Many efforts have been devoted to identifying movable parts from a single image~\cite{sun2023opdmulti,jiang2022opd} or from video sequences~\cite{qian2022understanding}. A large body of literature focuses on estimating joint parameters—such as rotation axes, pivot locations, and articulation angles—across different input modalities, including point clouds~\cite{fu2024capt,liu2023category,xu2022unsupervised,li2020category,yan2020rpm,jiang2022ditto,wang2019shape2motion}, RGB-D data~\cite{jain2022distributional,liu2022toward,che2024op,jain2021screwnet,abbatematteo2019learning,hu2017learning}, videos~\cite{liu2020nothing}, and 4D dynamic point clouds~\cite{liu2023building}. In addition to static scene understanding, active perception techniques have been introduced to enhance the precision of articulation estimation~\cite{zeng2024mars,yan2023interaction}. Furthermore, several works explore the temporal dynamics of articulated objects, addressing challenges such as continuous pose tracking and motion estimation~\cite{heppert2022category,weng2021captra}. Recently, vision-language models have been fine-tuned to support joint estimation tasks~\cite{huang2024a3vlm}. 

\vspace{-1em}
\subsection{Optimization-Based Articulated Object Reconstruction}
\vspace{-0.5em}

In addition to detecting and understanding 3D articulated objects, a substantial body of work focuses on reconstructing them from multi-view images. These methods typically follow a per-instance optimization paradigm, extending frameworks such as Neural Radiance Fields (NeRF)~~\cite{tseng2022cla, wei2022self, liu2023paris, weng2024neural, mu2021sdf, song2024reacto, wang2024sm, deng2024articulate, swaminathan2024leia} and 3D Gaussian Splatting~\cite{wu2025reartgs,liu2025building} to handle articulated objects. While effective at capturing object-specific geometry and appearance, these approaches are often slow due to their iterative optimization procedures. They also require densely sampled, posed images covering multiple articulation states, which are difficult to obtain in real-world settings, thereby limiting the scalability of these methods. Furthermore, they primarily rely on dense correspondence and lack cross-object priors, making it challenging to infer occluded or interior regions accurately.

\vspace{-1.2em}
\subsection{Feedforward Generation of 3D Articulated Objects}
\vspace{-0.3em}

Recent research has increasingly explored feedforward methods for the reconstruction and generation of 3D articulated objects, aiming to improve scalability and generalization. Some approaches focus on training 3D diffusion models specifically for articulated shapes~\cite{gao2024meshart,lei2023nap,luo2024physpart}, while others leverage fine-tuned vision-language models to predict joint parameters and part-level bounding boxes~\cite{le2024articulate}. Additionally, several methods adopt the paradigm of large-scale reconstruction model to synthesize novel views of articulated objects~\cite{gao2025partrm}. A diverse range of input modalities has been explored, including object categories and articulation graphs~\cite{liu2024cage}, single-view images~\cite{liu2024singapo,chen2024urdformer,dai2024automated,kawana2023detection}, multi-view inputs~\cite{mandi2024real2code,heppert2023carto,gadi2023rosi,zhang2021strobenet}, textual descriptions~\cite{su2024artformer}, and static 3D meshes~\cite{qiuarticulate}. Although these feedforward approaches generally enable faster inference, their performance is constrained by the limited availability of large-scale datasets for 3D articulated objects~\cite{xiang2020sapien}. Consequently, they often rely on coarse geometric approximations—such as bounding boxes, predefined templates, or retrieval from small-scale databases—to represent object parts, limiting their ability to model fine-grained and realistic geometry. Moreover, most existing methods focus primarily on geometry reconstruction and overlook texture modeling, resulting in outputs with reduced visual realism. While a few recent approaches~\cite{zhang2021strobenet,kawana2023detection,su2024artformer} employ implicit representations such as signed distance fields (SDFs) or occupancy fields to generate geometry, they still lack integrated texture synthesis and typically depend on brittle, multi-stage pipelines. In this work, we pursue a unified feedforward framework that jointly recovers detailed geometry, realistic textures, and accurate articulation structures for 3D articulated objects.

\vspace{-1.2em}
\subsection{Feedforward 3D Static Object Generation}
\vspace{-0.3em}
Open-world 3D generation of static objects has achieved notable progress. Recent approaches—such as CLAY~\cite{zhang2024clay}, Trellis~\cite{xiang2024structured}, and Hunyuan3D~\cite{zhao2025hunyuan3d}—pioneer native 3D diffusion models trained directly in the 3D domain. Enabled by large-scale datasets~\cite{deitke2023objaverse,deitke2023objaversexl} and advances in 3D representations, VAEs, and diffusion models, these methods produce accurate geometry, high-quality textures, and strong open-world generalization. However, extending this success to articulated objects by training native 3D diffusion models faces the critical challenge of limited data availability—current articulated datasets contain only a few thousand shapes~\cite{xiang2020sapien}, which is orders of magnitude fewer than their static counterparts.

In parallel, many works investigate feedforward reconstruction models. Some directly regress 3D shapes from single-view images~\cite{hong2023lrm,zou2024triplane,tochilkin2024triposr}, while others leverage sparse multi-view inputs~\cite{li2023instant3d,liu2023one,liu2024one,liu2024meshformer,wei2024meshlrm,wang2024crm,tang2024lgm,xu2024instantmesh,wu2024unique3d}. Though effective for static objects, these methods are rarely extended to articulated settings. Notably, a recent work, LVSM~\cite{jin2024lvsm}, demonstrates that a simple transformer model with minimal 3D inductive bias can achieve high-fidelity novel view synthesis for static objects. In this work, we build upon LVSM to support articulated objects—extending it to generate not only novel views under varying camera poses and joint states, but also explicit 3D textured meshes with disentangled parts and accurate articulation parameters.

\vspace{-0.7em}
\section{Method}
\vspace{-0.3em}

We introduce LARM, a unified framework for reconstructing 3D articulated objects from sparse-view RGB images captured at two distinct articulation states. The method enables articulation-aware novel view synthesis and supports explicit 3D reconstruction of both textured geometry and joint structure. We first describe the core model architecture, which encodes joint-aware input and target views to support both view synthesis and auxiliary predictions (Section~\ref{sec:larm_model}). Next, we show how the outputs of LARM are used to estimate joint parameters and reconstruct explicit 3D textured meshes (Section~\ref{sec:postprocessing}). Finally, we outline a two-stage training strategy and data augmentation techniques that facilitate network convergence and improve generalization (Section~\ref{sec:training_strategy}).

\vspace{-0.7em}
\subsection{LARM Model}
\vspace{-0.3em}
\label{sec:larm_model}
\noindent\textbf{Problem Formulation.} 
LARM takes as input $N \times 2$ sparse-view images of the object captured at two different joint states, along with known camera extrinsics and intrinsics. The input can be represented as:
\begin{equation}
\small
\left\{\left(\theta_i, \mathbf{I}_{ij}, \mathbf{E}_{ij}, \mathbf{K}_{ij}\right) ,\middle|, i = 1,2;\ j = 1,\dots,N \right\},
\end{equation}
where $\theta_i$ denotes the articulation state, $\mathbf{I}_{ij}$ is the $j$-th input image at state $\theta_i$, and $\mathbf{E}_{ij}$ and $\mathbf{K}_{ij}$ are the corresponding camera extrinsics and intrinsics. Given a target joint state $\theta_t$ and a novel camera pose defined by extrinsics $\mathbf{E}_t$ and intrinsics $\mathbf{K}_t$, LARM synthesizes the corresponding target-view image $\mathbf{I}_t$, along with optional outputs (e.g., depth maps or segmentation masks), which can be used for explicit 3D reconstruction.

While the two joint states, $\theta_1$ and $\theta_2$, are used as input, they are not required to tie to specific physically interpretable quantities such as degrees or distances. For instance, one can simply assign the ``rest'' state as $\theta_1 = \mathbf{0}$ and the ``maximum'' joint state as $\theta_2 = \mathbf{1}$, regardless of whether the actual articulation corresponds to 80 degrees or 0.3 meters. This relative scale avoids the need to annotate exact physical quantities for the input images. In this paper, we mainly focus on the cleaner and more useful setting where inputs only involve a single joint, such that $\theta_i$ is a scalar. However, in Section~\ref{sec:multi-part}, we show that extending to multi-part inference, such as reconstructing a cabinet with multiple doors and drawers, is straight-forward and only requires minimal modifications to our pipeline. 

\begin{figure}
    \centering
    \includegraphics[width=\linewidth]{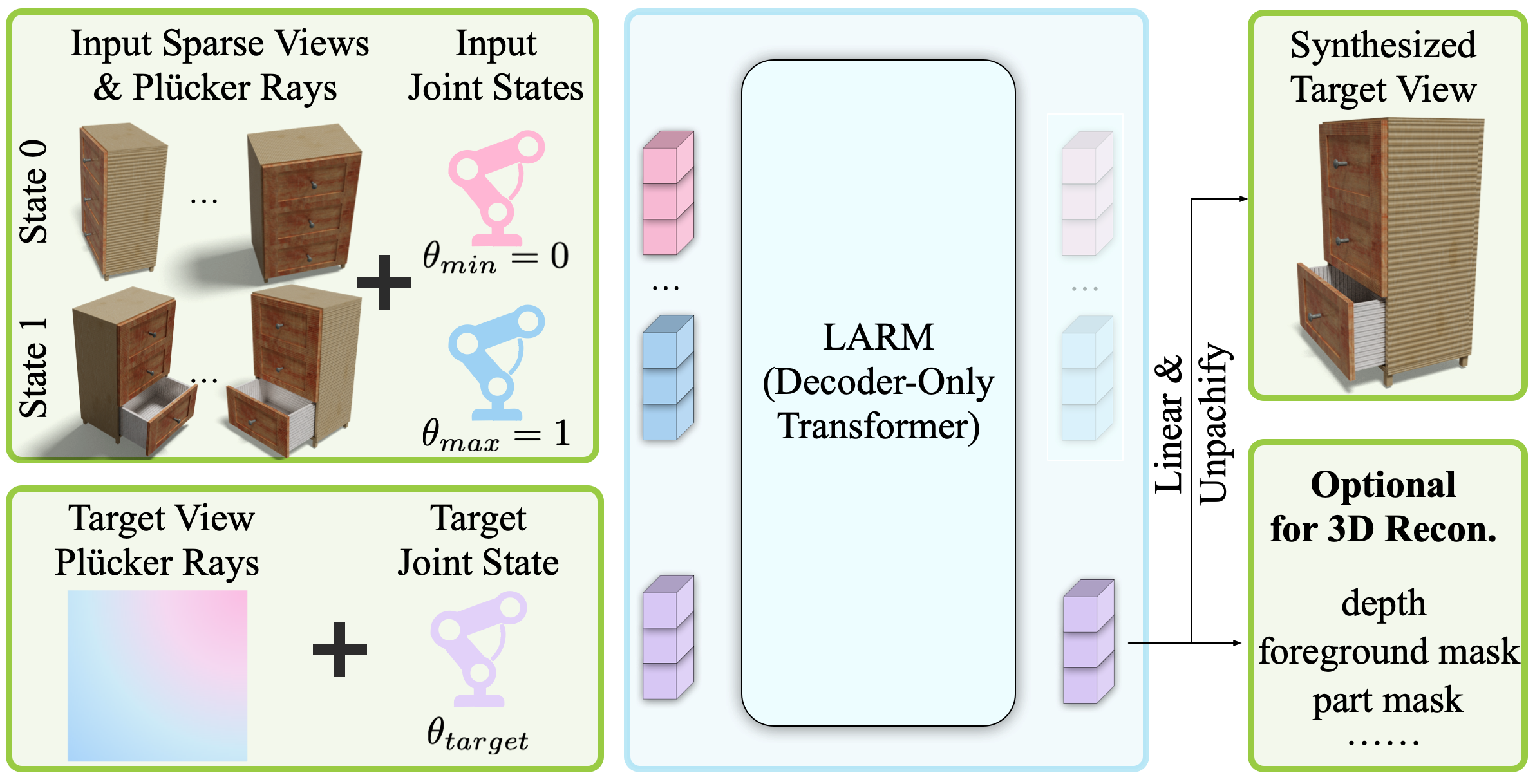}
    \vspace{-2.5em}
    \caption{\textbf{LARM Architecture.} LARM first patchifies the sparse, posed input images into tokens by concatenating the input RGB values, Plücker ray embeddings, and corresponding joint states. The target view to be synthesized is similarly represented by its Plücker ray embeddings and a target joint state, which are concatenated and tokenized. These input and target tokens are then fed into a decoder-only transformer model that predicts tokens used to regress the target view pixels. To enable explicit 3D reconstruction, LARM is also trained to produce additional outputs beyond RGB values, such as depth maps, foreground masks, and part masks.}
    \vspace{-0.5em}
    \label{fig:network}
\end{figure}

\noindent \textbf{Model Architecture} As shown in Figure~\ref{fig:network}, LARM is built upon LVSM's decoder-only architecture. For each input view $\mathbf{I}_{ij}$, we first compute a pixel-wise Plücker ray embedding $\mathbf{P}_{ij} \in \mathbb{R}^{H \times W \times 6}$ using the corresponding camera extrinsics and intrinsics, where $H$ and $W$ denote the height and width of the input images, respectively. Both the input image $\mathbf{I}_{ij}$ and its corresponding Plücker ray embedding $\mathbf{P}_{ij}$ are divided into patches of size $p \times p$, resulting in $\mathbf{I}_{ijk} \in \mathbb{R}^{3p^2}$ and $\mathbf{P}_{ijk} \in \mathbb{R}^{6p^2}$ for patch index $k = 1, \dots, HW / p^2$. For each patch, we concatenate the image patch, the Plücker ray embedding patch, and the corresponding joint state $\theta_i$, yielding a vector $\texttt{concat}([\mathbf{I}_{ijk}, \mathbf{P}_{ijk}, \theta_i]) \in \mathbb{R}^{9p^2+ 1}$.  This vector is then passed through a linear layer to produce an input patch token $\mathbf{x}_{ijk} \in \mathbb{R}^{d}$, where $d$ is the latent feature dimension. Target patch tokens are constructed similarly but only use the Plücker ray embedding patch from the target view $\mathbf{P}_{tk}$ and the target joint state $\theta_t$. These are concatenated and mapped through another linear layer to obtain the target-view patch tokens $\mathbf{q}_{tk} \in \mathbb{R}^{d}$.

We flatten and concatenate the token sequences from the input views, denoted as $x_1, \cdots, x_{l_x}$, and from the target view, denoted as $q_1, \cdots, q_{l_q}$, where $l_x = 2NHW / p^2$ and $l_q = HW / p^2$. The input tokens are fed into a decoder-only transformer model with multiple layers of self-attention. After applying the transformer layers, we obtain output tokens with the same sequence length as the input. Input view tokens $x_1, \cdots, x_{l_x}$ and target view tokens $q_1, \cdots, q_{l_q}$ are updated to $x'_1, \cdots, x'_{l_x}$ and $q'_1, \cdots, q'_{l_q}$, respectively. Each output target token $q'_i$ is passed through a linear layer to obtain a vector in $\mathbb{R}^{p^2c}$, where $c$ is the number of desired output channels. These vectors are reshaped into 2D patches, activated via a sigmoid function, and finally assembled to reconstruct the synthesized novel view $\hat{\mathbf{I}}_t$.

\noindent{\textbf{Auxiliary Output and Loss Functions} The original LVSM focuses solely on novel view synthesis of static objects and supervises training with the following loss:
\vspace{-0.3em}
\begin{equation}
\small
    \mathcal{L}_{RGB}=\operatorname{MSE}\left(\hat{\mathbf{I}}^t, \mathbf{I}^t\right)+\lambda_{perceptual} \cdot \operatorname{Perceptual}\left(\hat{\mathbf{I}}^t, \mathbf{I}^t\right),
\vspace{-0.3em}
\end{equation}

where $\lambda_{perceptual}$ is a weight that balances the rendering loss and perceptual loss.

In our setup, we can use the same loss to supervise LARM for modeling variations across both camera poses and joint states. However, in applications such as building digital twins in robotics, novel view synthesis alone is insufficient—we also require explicit 3D reconstruction of object geometry and articulation. To address this, we extend the model to additionally predict depth maps $\hat{\mathrm{D}}_t$ and $\hat{\mathrm{D}}_{ij}$, foreground masks $\hat{\mathrm{MF}}_t$ and $\hat{\mathrm{MF}}_{ij}$, and movable part masks $\hat{\mathrm{MP}}_t$ and $\hat{\mathrm{MP}}_{ij}$. Unlike LVSM, which discards input view tokens after transformer, LARM leverages both the updated input and target view tokens to generate these auxiliary outputs. Specifically, $\hat{\mathrm{D}}_t$, $\hat{\mathrm{MF}}_t$, and $\hat{\mathrm{MP}}_t$ are decoded from $q'_1, \dots, q'_{l_q}$, while $\hat{\mathrm{D}}_{ij}$, $\hat{\mathrm{MF}}_{ij}$, and $\hat{\mathrm{MP}}_{ij}$ are decoded from $x'_1, \dots, x'_{l_x}$, following the same decoding process used for synthesizing $\hat{\mathbf{I}}_t$. 

While only the auxiliary outputs corresponding to target views are used for explicit 3D reconstruction, the outputs from input views are leveraged during training to supervise the model and promote faster convergence. We supervise the full model using the following loss function:
\vspace{-0.3em}
\begin{equation}
\small
\mathcal{L} = \mathcal{L}_{RGB} + \lambda_{D} \mathcal{L}_{D} + \lambda_{MF}\mathcal{L}_{MF} + \lambda_{MP}\mathcal{L}_{MP},
\label{equ:loss}
\vspace{-0.3em}
\end{equation}
where $\mathcal{L}_{D}$ is the L1 loss applied to both predicted depth maps $\hat{\mathrm{D}}_{ij}$ and $\hat{\mathrm{D}}_t$. $\mathcal{L}_{MF}$ is the foreground mask loss for $\hat{\mathrm{MF}}_{ij}$ and $\hat{\mathrm{MF}}_t$, computed as a binary cross-entropy (BCE) loss with foreground and background pixels averaged separately. $\mathcal{L}_{MP}$ is the part mask loss for $\hat{\mathrm{MP}}_{ij}$ and $\hat{\mathrm{MP}}_t$, and it consists of two terms. The first term is computed similarly to $\mathcal{L}_{MF}$, using a separately averaged BCE loss over the full image. The second term accounts for cases where the part occupies a small region of the image: we crop the bounding box around the part and apply the BCE loss to the cropped region. The coefficients $\lambda{D}$, $\lambda_{MF}$, and $\lambda_{MP}$ are the corresponding weights used to balance the loss terms.

\vspace{-0.7em}
\subsection{Joint Estimation and Mesh Reconstruction}
\vspace{-0.3em}
\label{sec:postprocessing}
In this section, we describe how the outputs of the LARM model are used to estimate the underlying kinematic structure and reconstruct 3D meshes of articulated objects. 

\noindent\textbf{Joint Estimation.} We consider two common joint types: prismatic and revolute, and take the joint type as input, which we regard as a low-barrier requirement.\footnote{For instance, using a single image of each object, ChatGPT correctly identified the joint types for 88.89\% of our 144 test objects.}  For a revolute joint, we aim to estimate a pivot point $\mathbf{p} \in \mathbb{R}^3$ on the rotation axis, the axis direction $\mathbf{a} \in \mathbb{R}^3$, and a global joint scale factor $s$ that maps joint state differences to actual angular displacements. For a prismatic joint, we similarly estimate the axis direction $\mathbf{a}$ and the scale $s$.

\begin{figure}
    \centering
    \includegraphics[width=\linewidth]{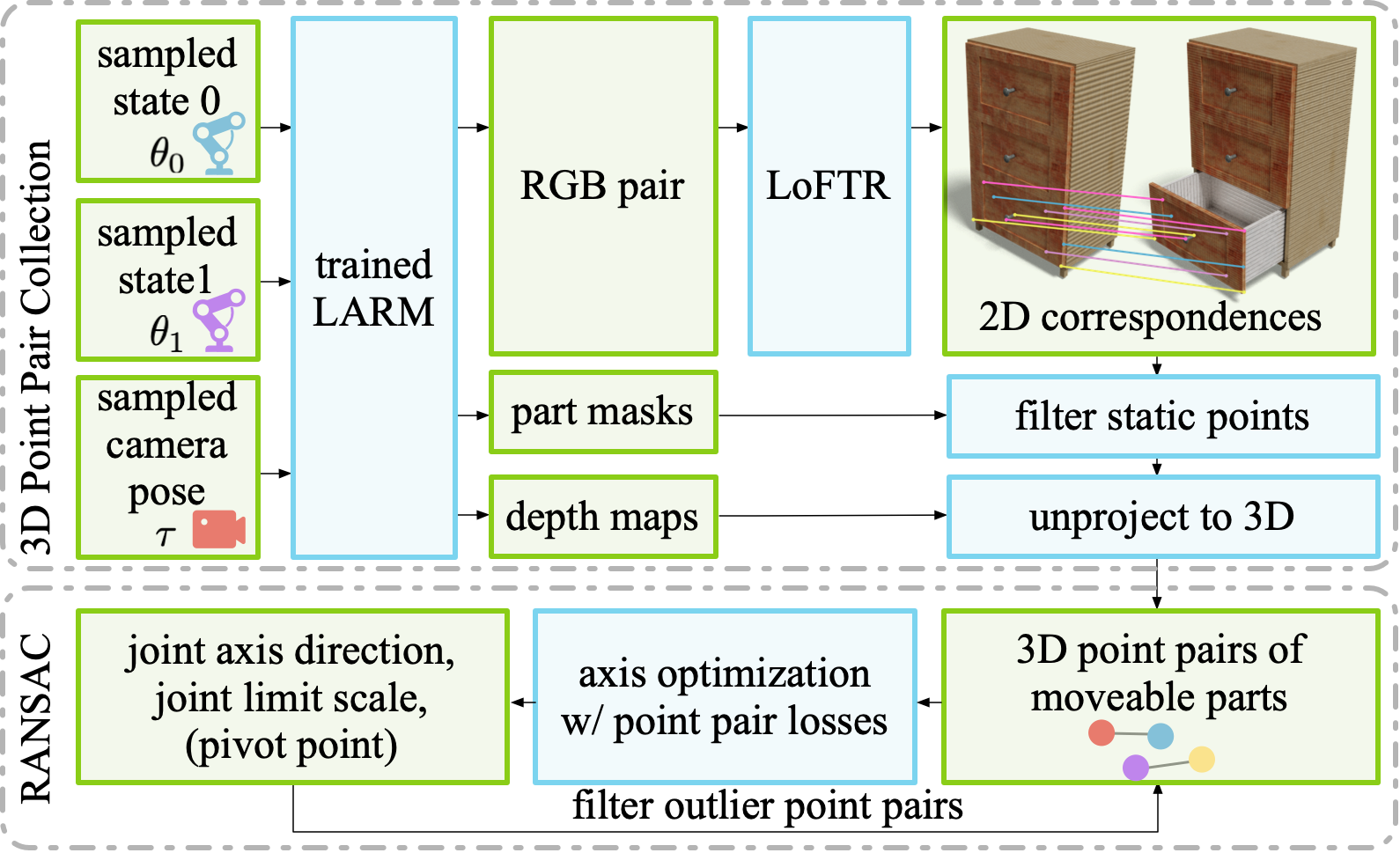}
    \vspace{-2em}
    \caption{\textbf{Joint Estimation.} To estimate explicit joint parameters using the LARM model, we first synthesize numerous image pairs with similar camera poses but different joint states. Next, we establish 2D pixel-wise correspondences for the movable part, which are then lifted to 3D. The joint parameters are optimized by minimizing the distances between the corresponding 3D point pairs under the estimated transformations. To enhance robustness, this optimization is integrated with RANSAC.}
    \label{fig:joint_estimation}
\end{figure}

To estimate these parameters, we query the LARM model with image pairs rendered under different joint states but with identical camera poses, as shown in Figure~\ref{fig:joint_estimation}. Each pair provides a view of the object at two articulation states, $\theta_u$ and $\theta_{v}$, while holding the viewpoint fixed. We then apply a 2D correspondence model (LoFTR~\cite{sun2021loftr}) to establish dense pixel-level matches between the two images. To ensure that the correspondences reflect true part motion, we retain only those matches located within the movable region, as defined by the predicted part masks $\hat{\mathrm{MP}}_u$ and $\hat{\mathrm{MP}}_{v}$. We filter correspondence pairs based on LoFTR confidence and remove those that are too close in pixel space.

Given predicted depth maps and camera poses, the surviving 2D correspondences are unprojected to produce sets of 3D point pairs. Each pair $(P_u^i, P_{v}^i)$ corresponds to a surface point observed at two joint states. The core idea is to model the transformation between these two points as a function of the joint parameters. For revolute joints, this transformation is a rotation around axis $\mathbf{a}$ passing through pivot $\mathbf{p}$ by an angle proportional to the joint state difference, $\Delta \theta = s(\theta_{u} - \theta_v)$. For prismatic joints, the transformation becomes a translation along $\mathbf{a}$ by a distance $\Delta d = s(\theta_{u} - \theta_v)$.

We then formulate an optimization problem to recover the joint parameters. The objective minimizes the 3D Euclidean distance between the transformed point $P_u^i$ and its paired point $P_{v}^i$:
\vspace{-0.5em}
\begin{equation}
    \small
    \mathcal{L}_{\text{joint}} = \sum_i \left\| T_{\theta_u \rightarrow \theta_{v}}(\mathbf{P}_u^i; \mathbf{a}, \mathbf{p}, s) - \mathbf{P}_{v}^i \right\|_2^2,
\vspace{-1em}
\end{equation}
where $T_{\theta_u \rightarrow \theta_{v}}$ denotes the appropriate joint-induced transformation (either rotational or translational). We optimize this objective using gradient-based methods to jointly solve for $\mathbf{a}$, $\mathbf{p}$, and $s$. RANSAC~\cite{fischler1981ransac} is applied to eliminate outliers in joint estimation.

\noindent\textbf{Mesh Reconstruction} As shown in Figure~\ref{fig:TSDF}, we sample a range of joint states and camera poses and use LARM to generate synthetic outputs for each configuration. The predicted depth maps are unprojected into 3D space using the known camera parameters, resulting in colored point clouds from multiple views. Each 3D point is associated with both a color (from the synthesized image) and a part label—either body or movable—based on the predicted part and foreground masks. This labeling enables the separation of raw point clouds into distinct semantic groups.

To reconstruct surface meshes, we first fuse the multi-view point clouds and then put the result into an off-the-shelf point cloud-to-mesh tool~\cite{peng2021shape} for mesh reconstruction. Importantly, we perform this reconstruction process separately for the movable part and the main body, resulting in two disjoint meshes that can articulate independently based on the estimated joint parameters. Our method recovers a single set of canonical meshes and estimates joint parameters by leveraging multi-state observations. To obtain meshes under a specific articulation state, we simply transform the canonical meshes using the inferred joint parameters instead of reconstructing meshes again.

\begin{figure}
    \centering
    \includegraphics[width=\linewidth]{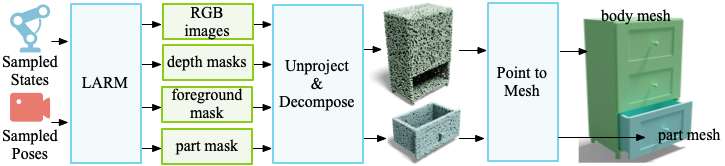}
    \vspace{-2em}
    \caption{\textbf{Mesh Reconstruction.} To extract an explicit mesh using the LARM model, we first synthesize multiple views from different camera poses. These views are lifted to 3D using the predicted depth and segmented based on the predicted masks, forming two separate colored point clouds. These point clouds are then fed into off-the-shelf point cloud-to-mesh tools~\cite{peng2021shape} for explicit mesh reconstruction.}
    \label{fig:TSDF}
\end{figure}

\vspace{-0.7em}
\subsection{Data and Training Strategy} 
\vspace{-0.3em}

\label{sec:training_strategy}
A major challenge in modeling articulated objects lies in the scarcity of high-quality datasets. To overcome this limitation and ensure both generalization and task-specific specialization, we adopt a two-stage training strategy: large-scale pretraining followed by task-specific finetuning.

\noindent\textbf{Pretraining on Static Objects} We begin by pretraining the LARM model on the Objaverse~\cite{deitke2023objaverse} dataset, which contains a diverse set of static 3D objects without articulation. During this stage, the model is trained purely for novel view synthesis using only RGB supervision $\mathcal{L}_{RGB}$. This stage equips the model with strong view synthesis capabilities and robust representations of object geometry and appearance. To further improve efficiency and performance, we adopt a coarse-to-fine training schedule: the model is first trained at a resolution of 256×256 and then refined at 512×512 resolution.

\noindent\textbf{Finetuning on Articulated Objects} After pretraining, we finetune the model on articulated objects from the PartNet-Mobility dataset, which includes detailed annotations of part articulation. In this stage, we activate the full prediction heads (for RGB, depth, masks, etc.) and supervise the model with the complete loss function introduced earlier. The linear output heads for RGB, depth, and masks are initialized by copying and repeating the pretrained RGB head weights, allowing the model to produce reasonable outputs from the very beginning of finetuning. This warm-start helps stabilize training and accelerates convergence.

\noindent\textbf{Data Augmentation} To enhance robustness and encourage generalization across unseen articulation patterns, we apply a series of augmentation strategies to the articulated dataset during finetuning. These include random scaling of objects along different axes to simulate variations in shape proportions, as well as random texture augmentation to diversify appearance.

\vspace{-0.7em}
\section{Experiment}

\vspace{-0.3em}
\subsection{Implementation and Evaluation Details}
\vspace{-0.3em}
In our experiments, we set $N = 3$, meaning that we use a total of 6 input images captured from two different joint states. The LARM model consists of 12 layers of self-attention with a bidirectional all-ones attention mask and a hidden dimension of 768. We first pretrain LARM on the novel view synthesis task using 3D static objects. Specifically, we train on a subset of 760k meshes from the Objaverse dataset~\cite{deitke2023objaverse}. For each mesh, we randomly sample 32 images to serve as the input and target view sources. The model is pretrained using 4 L40S GPUs for 3 days. After pretraining, we finetune the LARM model on articulated objects from the PartNet-Mobility dataset~\cite{xiang2020sapien}. We focus on six categories: StorageFurniture, Microwave, Refrigerator, Safe, TrashCan, and Table, with a total of 572 meshes (1,045 joints). We randomly hold out 116 shapes (with 293 joints in total) as test samples. To augment the training data, we apply random scaling and texture augmentation, where texture maps are randomly sampled from a database to replace the originals. For each joint movement, we generate eight variations (an object with two joints is augmented 16 times). For each variant, we render 32 views from randomly sampled camera poses and joint states, resulting in over 300,000 rendered views, which serve as the input and target images for training. Note that the train–validation–test splits are determined by objects rather than by images, which means all views of a test object are unseen during training.  We finetune the LARM model using the loss described in Equation~\ref{equ:loss}, employing 32 H100 GPUs for 5 days. The loss weights $\lambda_{\text{perceptual}}$, $\lambda_{D}$, $\lambda_{MF}$, and $\lambda_{MP}$ are set to 0.1, 2.0, 1.0, 1.0 respectively. With the inferred novel views, mesh reconstruction and joint estimation for each object take approximately 90 seconds.

\subsection{Novel View and State Synthesis} 

\begin{table}[t]
  \centering
  \small
  \setlength{\tabcolsep}{1pt}
  \caption{\textbf{Novel View and State Synthesis Results.} We report PSNR scores on our testing categories to reflect novel view synthesis quality, compared against Paris as the baseline method.}
  \vspace{-1em}
    \begin{tabular}{c|ccccccc|c}
    \toprule
          & Input & \textcolor[rgb]{ .114,  .11,  .114}{Storage.} & \textcolor[rgb]{ .114,  .11,  .114}{Micro.} & \multicolumn{1}{l}{Refrig.} & Safe  & TrashCan & Table & Average \\
    \midrule
    Paris & dense view & 21.54 & 22.63 & 21.92 & 19.24 & 20.90 & 21.35 & 21.26 \\
    Ours  & sparse view & \textbf{33.06} & \textbf{30.42} & \textbf{31.25} & \textbf{30.06} & \textbf{29.89} & \textbf{29.89} & \textbf{30.76} \\
    \bottomrule
    \end{tabular}%
  \label{tab:nvs}%
  \vspace{-1em}
\end{table}%
\begin{table}[t]
\centering
\small
\setlength{\tabcolsep}{1pt}
\caption{\textbf{Cross-State Consistency.} We generate videos of state-interpolation predictions and evaluate their temporal consistency by computing the Temporal LPIPS metric. This metric measures the LPIPS between adjacent frames and is computed over 1172 videos (25 frames each).}
\vspace{-1em}
\begin{tabular}{c|ccc}
\toprule
& Paris & Ours & GT \\
\midrule
Temporal LPIPS $\downarrow$ & 0.013 & \textbf{0.006} & \textbf{0.006} \\
\bottomrule
\end{tabular}
\label{tab:video_consistency}
\end{table}
\begin{table*}[t]
  \centering
  \small
  \setlength{\tabcolsep}{3pt}
  \caption{\textbf{Comparison of Reconstructed Mesh Geometry.} We report Chamfer Distance and F-Score between the reconstructed meshes and ground-truth meshes from the PartNet-Mobility dataset. Each metric is first averaged across multiple joint states per object and then across different object instances. All baseline outputs are carefully aligned to match the format of PartNet-Mobility meshes, ensuring consistent and fair comparison. More detailed processing and setup of baseline methods can be found in the Appendix.}
  \vspace{-1em}
    \begin{tabular}{c|c|cccccccccccc|cc}
    \toprule
    \multirow{2}[2]{*}{} & \multirow{2}[2]{*}{Input} & \multicolumn{2}{c}{StorageFurniture} & \multicolumn{2}{c}{Microwave} & \multicolumn{2}{c}{Refrigerator} & \multicolumn{2}{c}{Safe} & \multicolumn{2}{c}{TrashCan} & \multicolumn{2}{c|}{Table} & \multicolumn{2}{c}{Average} \\
          &       & CD    & F-Score & CD    & F-Score & CD    & F-Score & CD    & F-Score & CD    & F-Score & CD    & F-Score & CD    & F-Score \\
    \midrule
    URDFormer & single view & 0.104 & 0.600 & 0.122 & 0.415 & 0.141 & 0.445 & 0.182 & 0.302 & 0.115 & 0.541 & 0.139 & 0.415 & 0.134 & 0.453 \\
    ArticulateAnything & single view & 0.092 & 0.647 & 0.089 & 0.687 & 0.111 & 0.598 & 0.165 & 0.348 & 0.101 & 0.638 & 0.092 & 0.637 & 0.108 & 0.592 \\
    Singapo & single view & 0.077 & 0.756 & 0.071 & 0.807 & 0.086 & 0.669 & 0.159 & 0.298 & 0.124 & 0.543 & 0.073 & 0.722 & 0.098 & 0.633 \\
    Paris & dense view & 0.035 & 0.942 & 0.032 & 0.925 & 0.046 & \textbf{0.913} & \textbf{0.046} & \textbf{0.847} & 0.048 & 0.893 & 0.072 & 0.955 & 0.046 & 0.913 \\
    Ours  & sparse view & \textbf{0.023} & \textbf{0.960} & \textbf{0.028} & \textbf{0.931} & \textbf{0.034} & 0.908 & \textbf{0.046} & 0.835 & \textbf{0.026} & \textbf{0.961} & \textbf{0.020} & \textbf{0.977} & \textbf{0.030} & \textbf{0.929} \\
    \bottomrule
    \end{tabular}%
    \vspace{-1.5em}
  \label{tab:geometry}%
\end{table*}%

\begin{table*}[t]
  \centering
  \small
  \setlength{\tabcolsep}{3pt}
  \caption{\textbf{Comparison of Reconstructed Mesh Appearance.} For each pair of reconstructed and ground-truth meshes, we render multi-view images and compute their PSNR and CLIP similarity. How we acquire textured mesh for comparison for each baseline method is detailed in the Appendix. All values are first averaged across multiple joint states and then across different shapes.
}
\vspace{-1.5em}
    \begin{tabular}{c|cccccccccccc|cc}
    \toprule
          & \multicolumn{2}{c}{\textcolor[rgb]{ .114,  .11,  .114}{StorageFurniture}} & \multicolumn{2}{c}{\textcolor[rgb]{ .114,  .11,  .114}{Microwave}} & \multicolumn{2}{c}{\textcolor[rgb]{ .114,  .11,  .114}{Refrigerator}} & \multicolumn{2}{c}{Safe} & \multicolumn{2}{c}{TrashCan} & \multicolumn{2}{c|}{Table} & \multicolumn{2}{c}{Average} \\
          & PSNR  & CLIP Sim & PSNR  & CLIP Sim & PSNR  & CLIP Sim & PSNR  & CLIP Sim & PSNR  & CLIP Sim & PSNR  & CLIP Sim & PSNR  & CLIP Sim \\
    \midrule
    URDFormer & 15.1  & 0.886 & 12.3  & 0.893 & 14.2  & 0.893 & 12.2  & 0.886 & 14.3  & 0.861 & 12.8  & 0.846 & 13.5  & 0.878 \\
    ArticulateAnything & 16.4  & \textbf{0.924} & 18.2  & \textbf{0.918} & 15.6  & 0.889 & 11.7  & 0.838 & 14.2  & 0.885 & 14.2  & 0.887 & 15.1  & 0.890 \\
    Singapo & 11.7  & 0.843 & 9.9   & 0.886 & 13.3  & 0.887 & 11.1  & 0.871 & 11.5  & 0.826 & 11.2  & 0.873 & 11.4  & 0.864 \\
    Ours  & \textbf{22.5} & 0.919 & \textbf{22.5} & 0.906 & \textbf{23.3} & \textbf{0.912} & \textbf{20.1} & \textbf{0.902} & \textbf{22.0} & \textbf{0.906} & \textbf{20.2} & \textbf{0.903} & \textbf{21.8} & \textbf{0.908} \\
    \bottomrule
    \end{tabular}%
    \vspace{-0.5em}
  \label{tab:appearance}%
\end{table*}%

\label{subsec:exp_nvs}
We compare our method with two recent state-of-the-art approaches, Paris~\cite{liu2023paris} and PartRM~\cite{gao2025partrm}, on the novel view and state synthesis task using our test set from the PartNet-Mobility dataset~\cite{xiang2020sapien}. Paris performs NeRF-based optimization for each shape and requires dense-view input for two joint states. PartRM takes a single view and a 2D drag prompt as input to synthesize a novel view after applying the drag operation. While both PartRM and our LARM model allow precise control over the joint state of the target view, we argue that the 2D drag prompt used by PartRM is less intuitive and less accurate for specifying the target joint state. For example, based on their released code, generating the drag prompt for evaluation on the PartNet-Mobility dataset requires access to the ground-truth 3D model. Although this setup is less practical, we still follow their protocol to generate the comparison results. Please refer to the supplementary materials for more details.

The qualitative results are shown in Figure~\ref{fig:nvs_sota}. As observed, PartRM fails in many cases, generating twisted shapes. Paris produces better results but still often generates blurry outputs and floaters, particularly around the movable parts, despite utilizing dense-view inputs. We speculate that its per-shape NeRF optimization lacks cross-shape priors, making it less effective at handling challenging articulated components. In contrast, LARM demonstrates a stronger ability to synthesize novel views with sharp details and accurate articulated motion. The quantitative comparison in Table~\ref{tab:nvs} further supports our observation, showing that LARM outperforms Paris by a large margin. 

To evaluate the cross-state consistency of the synthesized views, we ask the model to generate images of continuous state interpolation and then assess their temporal consistency using temporal LPIPS, which measures the LPIPS between adjacent frames. As shown in Table \ref{tab:video_consistency}, LARM maintains high temporal consistency, closely matching the ground truth and substantially outperforming the baseline Paris. We also provide qualitative examples in the supplementary materials.

\begin{figure*}
    \centering
    \includegraphics[width=\linewidth]{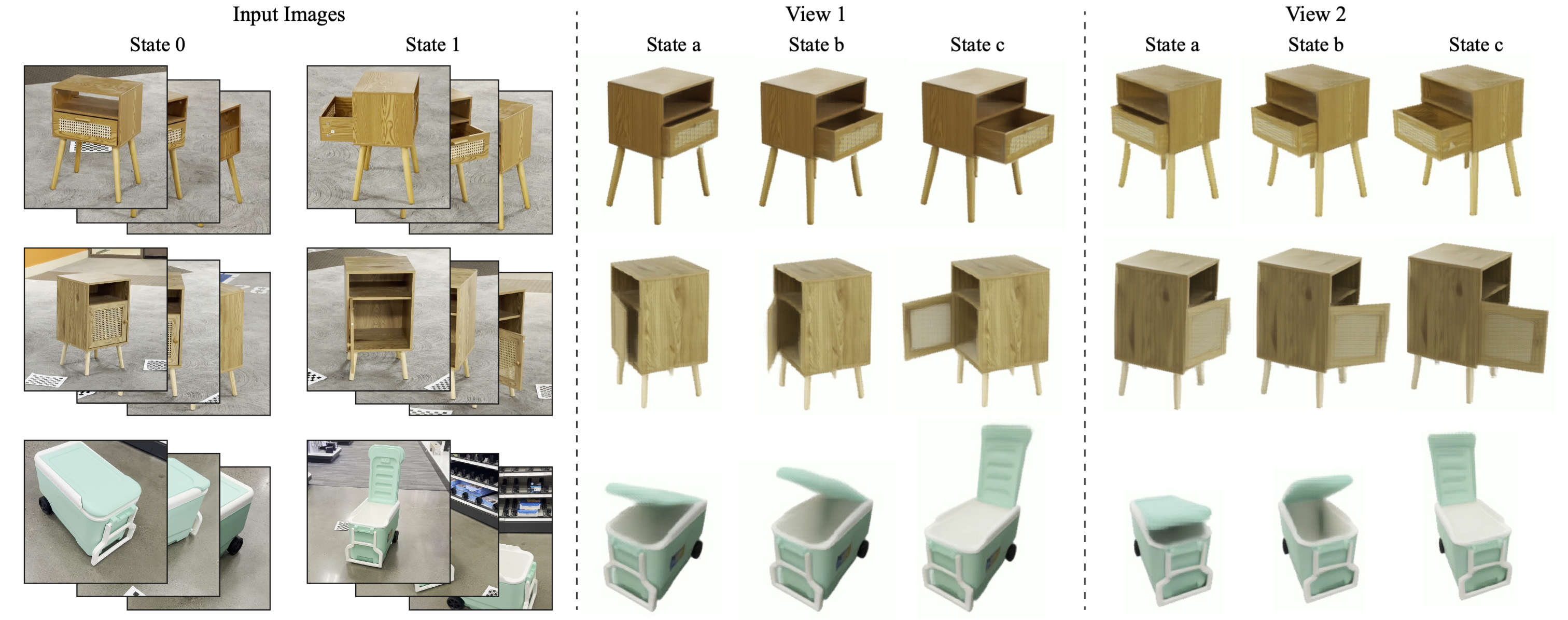}
    \vspace{-2.5em}
    \caption{\textbf{Real-world Demo.} We use an iPhone to capture sparse-view images of everyday articulated objects. The results demonstrate that LARM can effectively handle such inputs and predict accurate novel views across diverse camera poses and joint states.}
    \label{fig:real}
    \vspace{-1em}
\end{figure*}

\vspace{-0.3em}
\subsection{3D Articulated Object Reconstruction}
\vspace{-0.3em}

\begin{table*}[t]
  \centering
  \small
  \setlength{\tabcolsep}{3pt}
  \caption{\textbf{Multi-Part Reconstruction.} For each object, we sample five articulated instances by simultaneously sampling all joint states. When computing the metric, we first average over the five sampled articulated instances of each object and then average across objects.}
  \vspace{-1.5em}

  {
   \begin{tabular}{c|c|cccccccccccc|cc}
    \toprule
    \multirow{2}[2]{*}{} & \multirow{2}[2]{*}{Input} & \multicolumn{2}{c}{StorageFurniture} & \multicolumn{2}{c}{Microwave} & \multicolumn{2}{c}{Refrigerator} & \multicolumn{2}{c}{Safe} & \multicolumn{2}{c}{TrashCan} & \multicolumn{2}{c|}{Table} & \multicolumn{2}{c}{Average} \\
          &       & CD    & F-Score & CD    & F-Score & CD    & F-Score & CD    & F-Score & CD    & F-Score & CD    & F-Score & CD    & F-Score \\
    \midrule
    Singapo & single view & 0.104 & 0.603 & 0.131 & 0.504 & 0.113 & 0.509 & 0.158 & 0.381 & 0.149 & 0.471 & 0.120 & 0.517 & 0.114 & 0.556 \\
    Ours  & \(3K{+}3\) views & \textbf{0.031} & \textbf{0.940} & \textbf{0.018} & \textbf{0.975} & \textbf{0.066} & \textbf{0.729} & \textbf{0.040} & \textbf{0.947} & \textbf{0.025} & \textbf{0.970} & \textbf{0.051} & \textbf{0.841} & \textbf{0.038} & \textbf{0.905} \\
    \bottomrule
   \end{tabular}
  }
  \vspace{-1em}
  \label{tab:multilink}
\end{table*}
\begin{table}[t]
  \centering
  \scriptsize
  \setlength{\tabcolsep}{6pt}
  
  \caption{\textbf{Joint Estimation Accuracy.} We report the success rates of four metrics; see the text for their definitions and thresholds.}
  \vspace{-2em}

  {
   \begin{tabular}{c|cccc}
     \toprule
     Model & Axis Angle Error $\uparrow$ & Axis Origin Error $\uparrow$ & $M_r$ $\uparrow$ & $M_d$ $\uparrow$ \\
     \midrule
     Ours                & \textbf{81.2}\% & \textbf{81.9}\% & \textbf{87.0}\% & \textbf{84.6}\% \\
     Singapo             & 75.1\% & 80.2\% & 35.2\% & 50.9\% \\
     ArticulateAnything  & 63.8\% & 64.5\% & 32.8\% & 42.3\% \\
     Paris               & 20.1\% & 58.4\% & 42.0\% & 37.2\% \\
     URDFormer           & 51.9\% & 32.4\% & 27.7\% & 34.5\% \\
     \bottomrule
   \end{tabular}
  }
  \vspace{-2em}
  \label{tab:joint}
\end{table}

Beyond novel view and state synthesis, we compare our method with recent state-of-the-art approaches for 3D articulated object reconstruction: Articulate-Anything~\cite{le2024articulate}, Singapo~\cite{liu2024singapo}, URDFormer~\cite{chen2024urdformer}, and Paris~\cite{liu2023paris}. All methods except Paris operate on a single-view input. Articulate-Anything and Singapo construct articulated objects by retrieving parts and corresponding textures from the PartNet-Mobility dataset~\cite{xiang2020sapien}, whereas URDFormer predicts bounding boxes for individual parts, assembles template meshes, and projects the input image onto the mesh surface for texturing. During evaluation, we enumerate all joints of an object as target joints. For methods that predict multiple joints and links simultaneously, we compute the metric between their detected parts and the target part, selecting the one with the highest score. Please refer to the supplementary materials for more details.

We evaluate the reconstructed geometry (body and movable parts) based on both geometric and appearance accuracy. For each shape, we sample 5 joint states, generate corresponding meshes using the predicted joint parameters, and compute each metric for every state. The final results are reported as the average across all sampled states. Prior to evaluation, the predicted and ground-truth meshes are aligned using the method proposed in~\cite{liu2024one}. For geometric evaluation, we compute Chamfer Distance (CD)~\cite{fan2017point} and F-Score~\cite{wang2018pixel2mesh}, where CD is calculated using 100k sampled surface points, and F-Score is measured with a threshold of 0.05. For appearance evaluation, we render 8 random views for each state of both the predicted and ground-truth meshes, and compute the Peak Signal-to-Noise Ratio (PSNR) and CLIP similarity~\cite{radford2021learning}.

The qualitative results are shown in Figure~\ref{fig:mesh_sota}. We observe that URDFormer often produces overly simplistic geometry by compositing template board meshes and projecting the input images as textures, resulting in significant mismatches with the input prompts and ground-truth meshes. Articulate-Anything and Singapo achieve better results by retrieving part meshes from the PartNetMobility dataset. However, due to the limited scale of the retrieval database, they frequently capture only coarse geometry while failing to reproduce fine-grained geometric and texture details. In contrast, LARM, as a feedforward reconstruction model, faithfully reconstructs high-quality textured meshes that closely adhere to the input images. 

Table~\ref{tab:geometry} presents the quantitative evaluation of mesh geometry, where LARM outperforms all baselines across both metrics by a large margin. Notably, even though Paris utilizes dense-view inputs, it still performs slightly worse than our method, demonstrating the superiority of our feedforward model. Table~\ref{tab:appearance} shows the quantitative evaluation of mesh appearance. Since the released code of Paris did not export mesh textures, we excluded it from the appearance comparison and only compared against the other three baselines. As shown in the table, our method outperforms all baselines by a large margin, particularly in the PSNR metric. The poor performance of the baseline methods is primarily due to their reliance on shape retrieval or template mesh compositing strategies, which lead to significant discrepancies between the reconstructed and ground-truth meshes.

\vspace{-0.3em}
\subsection{Joint Parameter Estimation}
\vspace{-0.3em}

We follow Articulate-Anything~\cite{le2024articulate} to evaluate joint parameter estimation accuracy using four metrics: (1) Axis angle error, the angle between the predicted and ground-truth axis; (2) Axis origin error, the shortest distance between the two axes; (3) Motion range distance $M_r$, which measures how far the predicted motion range deviates from the ground truth; (4) Motion direction difference $M_d$, which checks whether the predicted motion points the same way as the true motion. We report the success rate for these four metrics using thresholds of 0.25 radians for axis angle error, 0.15 for axis origin error, and 0.3 for both $M_r$ and $M_d$. The results are shown in Table~\ref{tab:joint}, where our method outperforms all baselines on all metrics, demonstrating more accurate joint estimation.

\vspace{-0.3em}
\subsection{Ablation Study}
\vspace{-0.3em}

\begin{table}[t]
  \centering
  \scriptsize
  \setlength{\tabcolsep}{2.5pt}
  \caption{\textbf{Ablation Study.} Evaluated on the table category.}
  \vspace{-2em}
    \begin{tabular}{c|c|cccr|ccc}
    \toprule
    \multirow{3}[2]{*}{} & \multirow{3}[2]{*}{Model Variation} & \multicolumn{4}{c|}{LARM Prediction} & \multicolumn{3}{c}{Mesh Reconstruction} \\
          &       & Depth & NVS   & NVS   & \multicolumn{1}{c|}{mask} & \multirow{2}[1]{*}{CD} & \multirow{2}[1]{*}{F-Score} & Mesh \\
          &       & MAE   & PSNR  & LPIPS & \multicolumn{1}{c|}{mIoU} &       &       & PSNR \\
    \midrule
    a     & \textcolor[rgb]{ .114,  .11,  .114}{w/o static objects pretraining} & 0.034 & 24.397 & 0.096 & 0.9309 & \multicolumn{3}{c}{\multirow{2}{*}{\diagbox[innerwidth=3em]{}{} }}\\
    b     & \textcolor[rgb]{ .114,  .11,  .114}{w/o data augmentation} & 0.020 & 28.214 & 0.047 & 0.974 & \multicolumn{3}{c}{\multirow{2}[2]{*}{}} \\
    \midrule
    c     & \textcolor[rgb]{ .114,  .11,  .114}{reduced number of views (64->16)} & \multicolumn{4}{c|}{\multirow{2}{*}{\diagbox[innerwidth=4em]{}{} }} & 0.033 & 0.930 & 19.935 \\
    d     & \textcolor[rgb]{ .114,  .11,  .114}{reduced number of qpos (5->2)} & \multicolumn{4}{c|}{}         & 0.026 & 0.958 & 19.815 \\
    \midrule
    e     & full  & 0.014 & 29.888 & 0.039 & 0.9832 & 0.020 & 0.977 & 20.194 \\
    \bottomrule
    \end{tabular}%
  \label{tab:ablation}%
\end{table}%

We ablate our training and reconstruction strategies and report the results in Table~\ref{tab:ablation}.

\noindent\textbf{Pretraining on 3D Static Objects.} Due to the scarcity of 3D articulated objects, we first pretrain the model on a large-scale dataset of 3D static objects. This pretraining equips the model with fundamental knowledge of 3D object structures and the ability to synthesize novel views. When this pretraining is omitted and the model is trained from scratch on the limited set of articulated objects (row (a)), we observe a significant drop in performance, highlighting the importance of the pretraining stage.

\noindent\textbf{Data Augmentation.} To enrich the limited articulated objects, we apply augmentation of random scaling and the texture replacement. As shown in Row (b), this strategy effectively increases the data diversity and further improves the performance.

\noindent\textbf{Number of Views for Mesh Reconstruction.} LARM can synthesize novel views from arbitrary camera poses. For explicit mesh extraction, we use 64 views by default. In Row (c), we evaluate the effect of reducing this number to 16. While we observe a slight drop in performance, the results remain reasonable. Note that using fewer views facilitates faster reconstruction; however, when the number of views is further reduced to 4 or 8, holes occasionally appear in the reconstructed mesh.

\noindent \textbf{Number of Joint States for Joint Estimation.} LARM can synthesize outputs at novel joint states. To extract joint parameters (e.g., joint axis, pivot point), we query LARM at multiple joint states to construct correspondences and solve for the joint parameters. A greater number of joint states typically leads to more reliable correspondences and, consequently, more accurate joint estimation. In row (d), we reduce the default 5 sampled joint states to 2 and observe a drop in performance, highlighting the importance of LARM's ability to synthesize outputs at novel joint states.

\vspace{-0.3em}
\subsection{Real-world Evaluation}
\vspace{-0.3em}

Beyond evaluating our method on standard synthetic datasets, we also assess its performance on real-world, sparsely captured images. As shown in Figure~\ref{fig:real}, LARM can effectively handle iPhone-captured images of everyday articulated objects. This demonstrates the broad applicability of our method and its potential to support various real-world applications, such as building digital twins.

\vspace{-0.3em}
\subsection{Extension to Multi-Part Inference}
\vspace{-0.3em}

\label{sec:multi-part}
Extending our method to multi-part objects requires only an inference-time adaptation. Here we assume that the object has multiple parts, but each part possesses only a single degree of freedom, such as a cabinet with multiple drawers and doors. For $K$ movable parts, we use $3K + 3$ input images: three for the rest state (all parts closed) and three at the maximal articulation of each part. Each movable part is processed independently using its six images to infer the joint axis and reconstruct its part mesh, while the static body is recovered by fusing RGB-D point clouds and removing points belonging to movable parts via the inferred segmentation and depth.

We evaluate on the same test set under a multi-part setting, where the joint states of all joints are randomly sampled simultaneously and metrics are computed on the resulting articulated meshes. As shown in Table~\ref{tab:multilink}, our method remains effective in this multi-link scenario and outperforms the baseline Singapo. Note that we do not consider complex kinematic chains (e.g., a robot arm) in this work, which is consistent with most baselines that handle only single-DoF revolute or prismatic joints. This simplified setup covers a broad range of everyday articulated objects.
\section{Conclusion and Limitation}

We propose LARM, a novel feedforward model for reconstructing 3D articulated objects. Unlike recent approaches that rely on part retrieval or template meshes—which often result in significant mismatches compared to the input—LARM can reconstruct high-fidelity novel views and 3D textured meshes that closely match the input prompts. Looking ahead, although we have leveraged pretraining on 3D static objects, the performance and generalizability of the model are still limited by the scale of the articulated object dataset. Exploring better strategies for data synthesis or more advanced training techniques may further improve the model’s generalization and make it more widely applicable.
\clearpage
\begin{figure*}
    \centering
    \includegraphics[width=\linewidth]{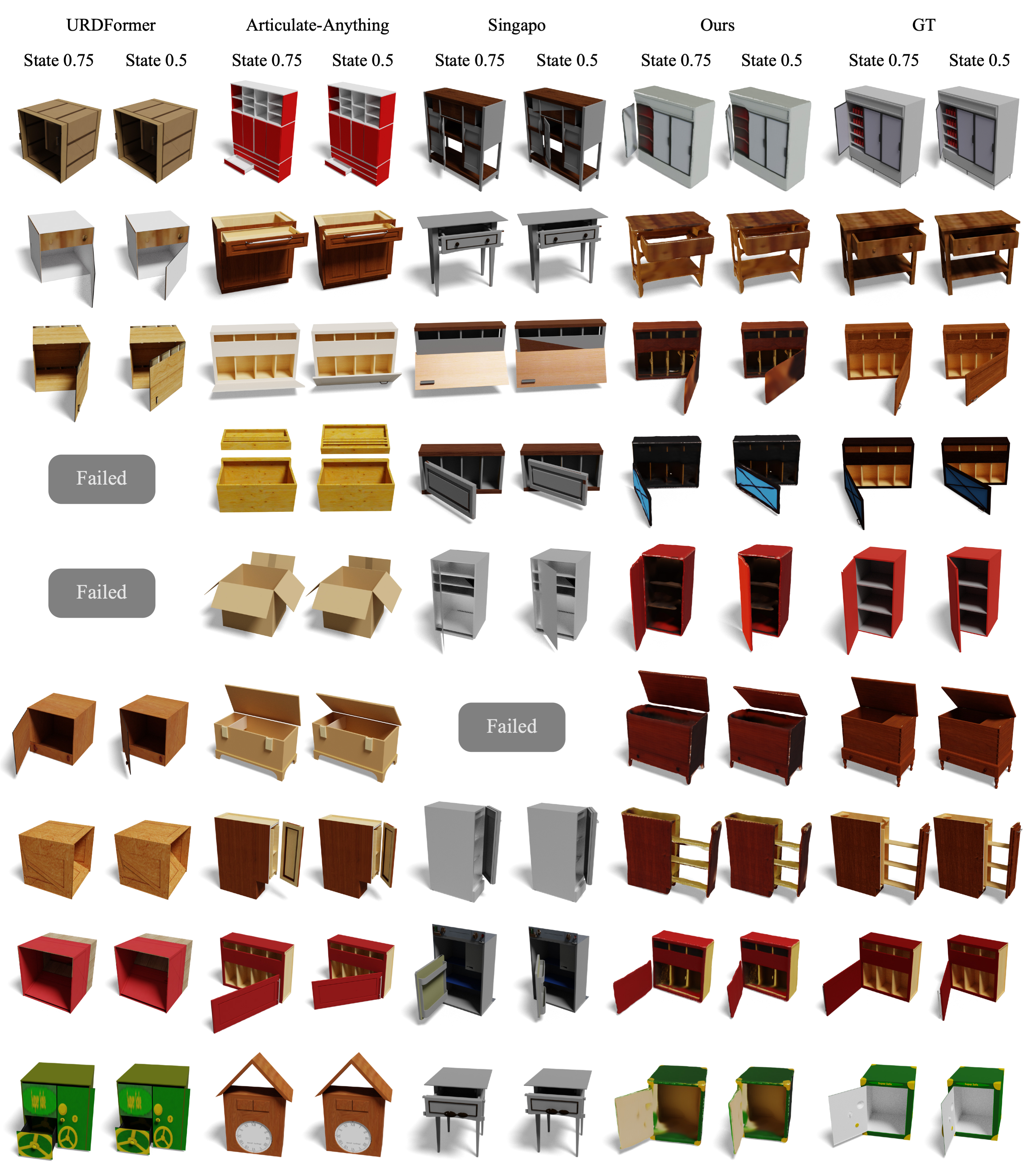}
    \vspace{-1em}
    \caption{\textbf{Comparison of 3D Articulated Object Reconstruction.} Note that methods such as URDFormer~\cite{chen2024urdformer} and Articulate-Anything~\cite{le2024articulate} rely on part retrieval for reconstruction, often resulting in significant mismatches in geometry and texture compared to the input prompt. In contrast, our LARM model faithfully reconstructs high-quality textured meshes that closely align with the input prompts.}
    \label{fig:mesh_sota}
\end{figure*} 

\begin{figure*}
    \centering
    \includegraphics[width=0.95\linewidth]{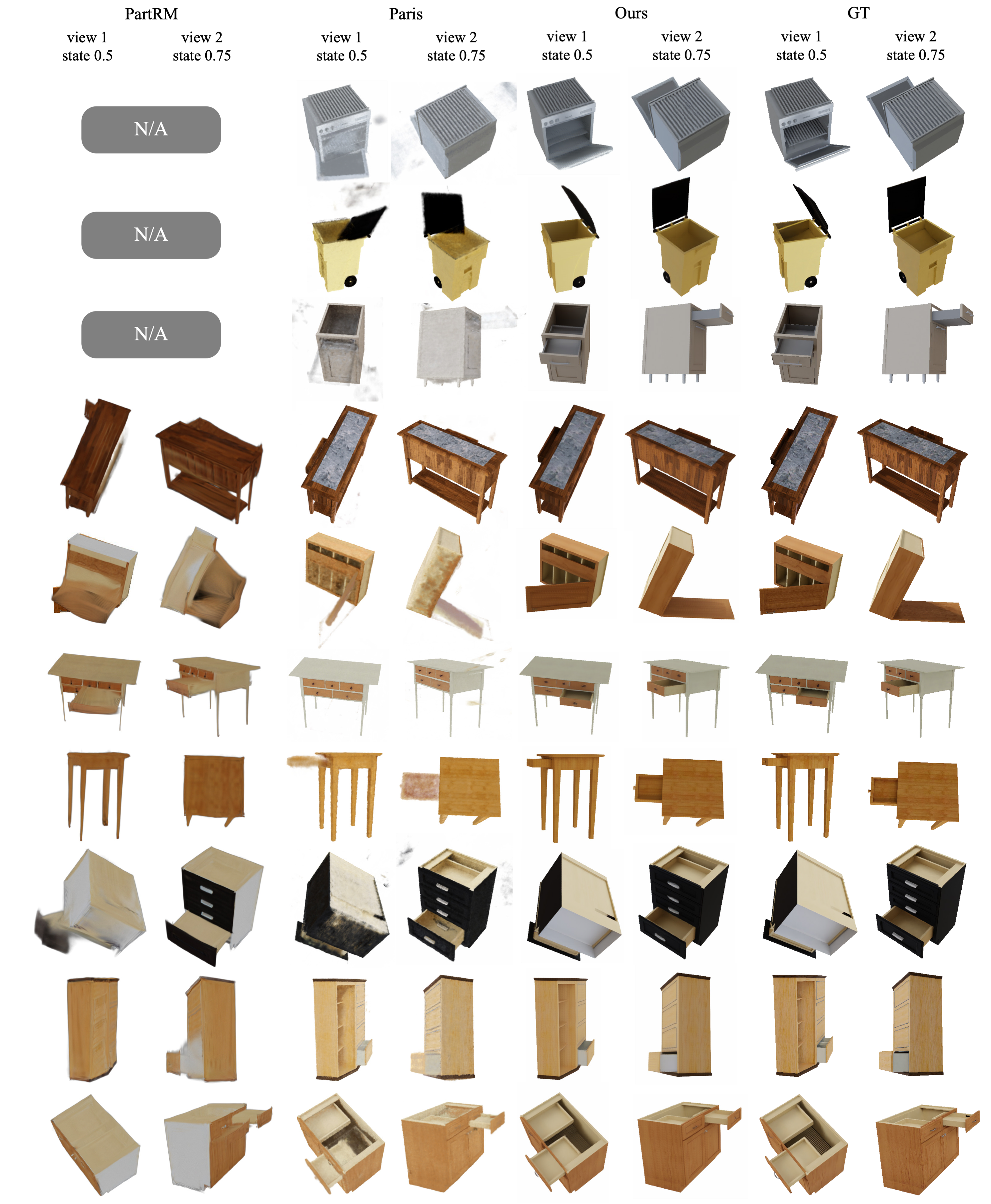}
    \caption{\textbf{Comparison of Novel View and State Synthesis between PartRM~\cite{gao2025partrm}, Paris~\cite{liu2023paris}, and our LARM.} For each shape, we showcase synthesized views at two novel joint states. For objects not included in the PartRM preprocessed datasets, we leave blank. Please refer to the appendix for details on PartRM evaluation.}
    \label{fig:nvs_sota}
\end{figure*}
\clearpage
\twocolumn[{%
  {\sffamily\bfseries\Huge Appendix}
  \par
  \vspace{2em}
}]
% Reset section counter for appendix
\setcounter{section}{0}
\renewcommand{\thesection}{A\arabic{section}} 
\section{Baseline Setup and Evaluation Details}

We evaluate each baseline under its as-published inference regime, using the released checkpoints when available or, for training-free methods, their provided prompts and inputs. Specifically, Articulate-Anything~\cite{le2024articulate} is a training-free, VLM-based method. For Paris~\cite{liu2023paris}, there is no dataset-level training; instead, it performs per-scene optimization from multi-view images. For URDFormer~\cite{chen2024urdformer} and Singapo~\cite{liu2024singapo}, we use their released checkpoints, which are pretrained on their respective training splits. We did not retrain these models, as they may require additional supervision beyond our rendered images. All baseline methods and our method are evaluated on the same test set.

\textbf{URDFormer} For evaluation of URDFormer\cite{chen2024urdformer}, we rendered and cropped a single front-view image from our Partnet-Mobility test set as input into the model, as a proxy to the real world image model input. We then used the provided fine-tuned Grounding DINO model for extracting predicted bounding boxes from the image. We did not manually adjust the output bounding boxes and instead directly used the predicted bounding boxes in the pipeline for articulation prediction. URDFormer focuses on predicting articulation and joint information over producing realistic meshes and provides predefined mesh templates for visualization of output 3D meshes. In our evaluation, we used these predefined mesh templates, combined with their extracted texture from the input image, as the base meshes for rendering and computing various geometric and appearance metrics. 

\textbf{Articulate-Anything} For evaluating Articulate-Anything\cite{le2024articulate}, we used its visual-input-based inference mode, providing single-view rendered images from our test set as input. We employed Google Gemini 1.5 Flash as the language model. Since the visual inference mode of Articulate-Anything is limited to retrieving entire objects from the PartNet-Mobility dataset, we excluded the target object itself from the candidate pool during inference to prevent exact matches and ensure a fair evaluation.

\textbf{Singapo} For evaluation of Singapo\cite{liu2024singapo}, similarly, we provide a single-view rendering of our test set object as input. We use GPT-4o as the language model for acquiring input structural graphs from the input image. Singapo retrieves part meshes from partnet mobility and does not attempt to match the input image in texture. However, as a proxy, we instead enforces retrieval of texture along with any desired part mesh. The appearance and rendering metrics were computed with such retrieved textures. 

\textbf{Paris} To construct the dense view input for Paris\cite{liu2023paris}, we rendered  100 training views for each input object joint, under the rest state and the maximum state respectively, ensuring same lighting condition as the target view renderings. Aside from novel view synthesis, we also used the extracted mesh for computation of geometry metrics.

For all of these above baseline methods, the resulting meshes are centered and normalized in the same way as the partnet mobility object meshes to ensure alignment and reasonable comparison of metrics. 

\textbf{PartRM} Because data preprocessing for PartRM\cite{gao2025partrm} is particularly involved, we followed the exact evaluation pipeline from their released code— including drag prompts computed from the ground-truth meshes. To give PartRM the most favorable conditions, we bypassed their original Zero123++ stage (which infers multi-view images from a single view) and instead fed their reconstruction model the corresponding ground-truth images directly. Likewise, to eliminate any domain gap from differing camera intrinsics, we used the authors’ own preprocessed PartNet-Mobility as input and applied their pre-trained weights without alteration. Please note that these comparisons are strictly qualitative: our rendering parameters differ from theirs, and their preprocessed dataset omits several object categories. In those cases, we have left the corresponding slots blank in our visualizations.

\begin{table}[t]
  \centering
  \small
  \setlength{\tabcolsep}{3pt}
  \caption{\textbf{Category Generalization to an Unseen Class (Oven).} LARM is trained without any oven instances and evaluated on the Oven category at test time. Each metric is first averaged across multiple joint states per object and then across object instances. Lower values indicate better performance for CD, whereas higher values are better for the other metrics.}
  \vspace{-1em}

  {
   \begin{tabular}{c|ccccc}
     \toprule
     \multirow{2}[2]{*}{Model} & \multicolumn{5}{c}{Oven} \\
      & CD $\downarrow$ & F\mbox{-}Score $\uparrow$ & Render PSNR $\uparrow$ & CLIP $\uparrow$ & NVS PSNR $\uparrow$ \\
     \midrule
     URDFormer     & 0.1335 & 0.4040 & 13.1878 & 0.8709 & \NAcell \\
     ArtAnything   &  0.0632 & 0.8146 & 14.4753 & 0.8818 & \NAcell \\
     Singapo       & 0.0714 & 0.7522 & 11.5511 & 0.8693 & \NAcell \\
     Paris         &  0.0343 & 0.9178 & \NAcell & \NAcell & 19.9293 \\
     Ours   & \textbf{0.0283} & \textbf{0.9272} & \textbf{20.5066} & \textbf{0.9671} & \textbf{28.0743} \\
     \bottomrule
   \end{tabular}
  }
  \vspace{-1.2em}
  \label{tab:oven}
\end{table}

\section{Qualitative Examples of Cross-State Consistency}
In Figure~\ref{fig:video_consistency}, we showcase qualitative novel view synthesis results of articulation state interpolation, where our method exhibits strong cross-state consistency.

\section{Category Generalization}

We acknowledge that our method’s generalizability is still limited by the scale of training data—a common challenge for feedforward approaches. However, unlike many baselines that rely on category-specific part structure templates or part retrieval databases, our method is both template-free and database-free, offering greater potential for generalization.

To evaluate the generalizability of LARM to unseen categories, we exclude \emph{Oven} from training and assess performance on it at test time. As shown in Table~\ref{tab:oven}, our model outperforms baselines trained with oven data across geometry (CD, F-Score), appearance (Render PSNR, CLIP), and novel-view synthesis (NVS PSNR), demonstrating stronger category-level generalization. While LARM generalizes well to categories similar to the training set, categories that differ substantially from the training distribution (e.g., scissors) remain challenging.

\section{Discussion of Failure Cases}
We observe failures when (i) test shapes or kinematics deviate significantly from the training data (e.g., scissors), leading to inaccurate and blurry novel-view synthesis results, particularly in the movable part regions; (ii) sparse or self-occluded inputs undersample movable parts, causing unstable correspondences; and (iii) textureless or reflective regions impair depth prediction and 2D point matching. Figure~\ref{fig:failure} shows sample failure cases as discussed above.

\begin{figure*}[t]
  \centering
  \includegraphics[width=0.95\linewidth]{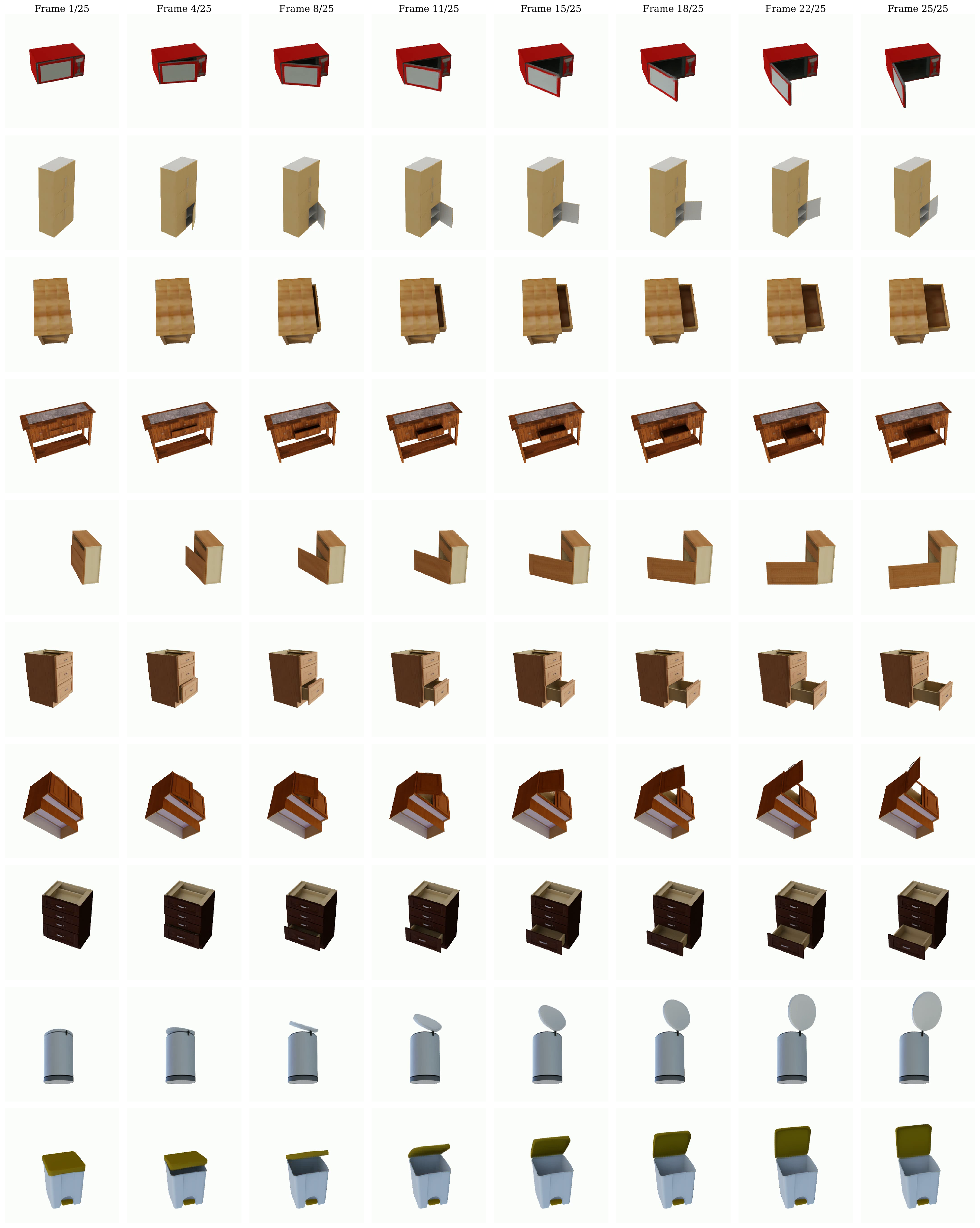}
  \vspace{-1.0em}
  \caption{\textbf{Qualitative Example of Cross-State Consistency.} For each example, we visualize NVS results across eight evenly sampled articulated states.}
  \label{fig:video_consistency}
  \vspace{-0.6em}
\end{figure*}

\begin{figure*}[t]
  \centering
  \includegraphics[width=0.85\linewidth]{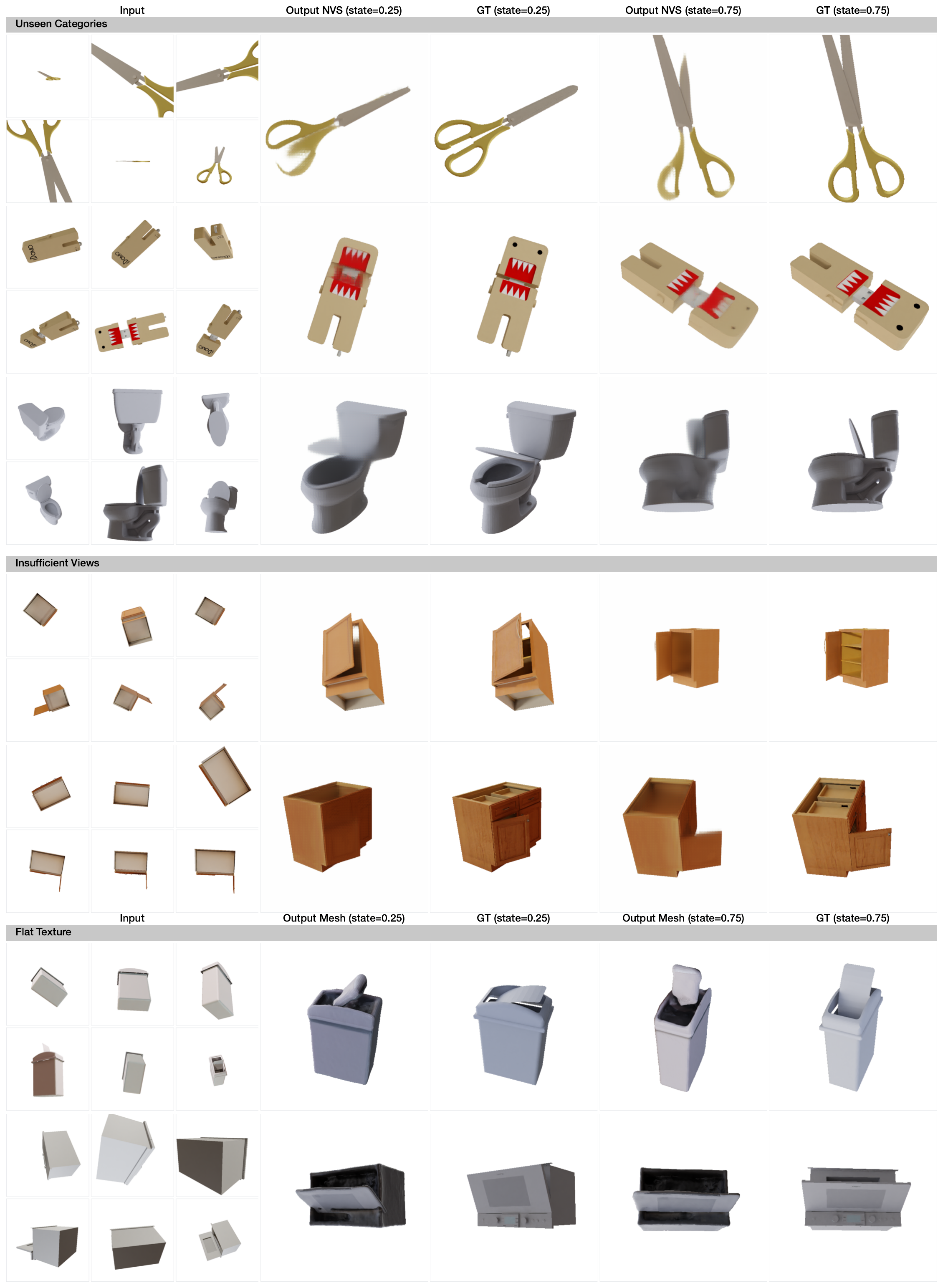}
  \vspace{-1.0em}
  \caption{\textbf{Failure Cases.} Here we show failure cases of LARM. Failure cases occur under three main categories: (i) test shapes are significantly different from the training data (e.g., scissors, rows 1 - 3); (ii) sparse or self-occluded inputs (rows 4-5); and (iii) large textureless or reflective regions on target objects (rows 6-7).}
  \label{fig:failure}
  \vspace{-0.6em}
\end{figure*}

\clearpage
\bibliographystyle{ACM-Reference-Format}
\bibliography{bib}

\end{document}